\documentclass[runningheads]{llncs}
\usepackage[T1]{fontenc}
\usepackage[dvipsnames]{xcolor}
\usepackage{graphicx,verbatim}
\usepackage[pagebackref=true,breaklinks=true,colorlinks,bookmarks=false]{hyperref}
\usepackage{amsmath}
\usepackage{amssymb}
\usepackage{subcaption}
\usepackage{enumitem}
\usepackage{esvect}

\newcommand{\ProposedMethod}{QwenVLConnector}
\begin{document}
\title{\ProposedMethod{}: A Fast, Unified Medical VLM Chatbot for Fine-Grained Clinical Perception and Text Generation}

\titlerunning{\ProposedMethod{}}
%
\begin{comment}  %% Removed for anonymized MICCAI 2025 submission
\author{First Author\inst{1}\orcidID{0000-1111-2222-3333} \and
Second Author\inst{2,3}\orcidID{1111-2222-3333-4444} \and
Third Author\inst{3}\orcidID{2222--3333-4444-5555}}
%
\authorrunning{F. Author et al.}
% First names are abbreviated in the running head.
% If there are more than two authors, 'et al.' is used.
%
\institute{Princeton University, Princeton NJ 08544, USA \and
Springer Heidelberg, Tiergartenstr. 17, 69121 Heidelberg, Germany
\email{lncs@springer.com}\\
\url{http://www.springer.com/gp/computer-science/lncs} \and
ABC Institute, Rupert-Karls-University Heidelberg, Heidelberg, Germany\\
\email{\{abc,lncs\}@uni-heidelberg.de}}

\end{comment}

% \author{Anonymized Authors}  %% Added for anonymized MICCAI 2025 submission
% \authorrunning{Anonymized Author et al.}
% \institute{Anonymized Affiliations \\
%     \email{email@anonymized.com}}

\author{
Le Thien Phuc Nguyen\inst{1,3}$^*$\and
Hoang-Thien Nguyen\inst{2}$^*$ \and
Thanh-Huy Nguyen\inst{6} \and
Gia Minh Hoang\inst{4}\and
Mai-Anh Vu\inst{5} \and 
Ulas Bagci\inst{6}
}
\renewcommand\thefootnote{\fnsymbol{footnote}}
\footnotetext[1]{These authors contributed equally.}

\authorrunning{Le Thien Phuc Nguyen \& Hoang-Thien Nguyen et al.}

%
% \authorrunning{F. Author et al.}
% First names are abbreviated in the running head.
% If there are more than two authors, 'et al.' is used.
%
\institute{
University of Wisconsin - Madison, USA \and
Carnegie Mellon University, USA \\
hoangthn@andrew.cmu.edu \and
University of North Carolina - Chapel Hill \\
\email{tphuc@cs.unc.edu} \and
Mayo Clinic, College of Medicine and Science, Scottsdale, USA \and 
University of Houston, USA  \and %mvu9@cougarnet.uh.edu 
Northwestern University, USA \\ \email{thanhnguyen2031@u.northwestern.edu,ulas.bagci@northwestern.edu}
}

\maketitle              % typeset the header of the contribution
\begin{abstract}
Most medical vision–language models (VLMs) excel at open-ended report generation and VQA but provide limited support for structured, fine-grained clinical perception within a unified interface. We present QwenVLConnector, a Qwen2.5-VL–based medical chatbot that unifies classification, multi-label classification, textualized detection, counting, regression, and free-form report generation under a single next-token objective. Our key component is a lightweight dense multi-layer Connector that aggregates low- and high-level visual features, aligns them through the pretrained vision Merger, and fuses them with the final visual representation without increasing sequence length. This design enriches visual tokens with complementary spatial and semantic cues while preserving efficiency. On FLARE-2D, QwenVLConnector improves detection F1 from 0.55 to 0.85, raises single-label classification from 0.37 to 0.51, and boosts report-generation GREEN by up to 18.3 points over the Qwen2.5-VL baseline. We further explore multimodal in-context learning for report generation, showing additional improvements without updating model parameters. Overall, QwenVLConnector offers a unified and efficient framework for combining structured medical perception with open-ended clinical text generation. Our code is available at \url{https://github.com/plnguyen2908/QwenConnector}. 
\end{abstract}

\section{Introduction}

Medical vision–language chatbots have progressed rapidly from task-specific tools to assistants that can describe, reason about, and converse over clinical imagery. Systems such as LLaVA-Med~\cite{NEURIPS2023_5abcdf8e} and HuatuoGPT-Vision~\cite{chen2024huatuogptvisioninjectingmedicalvisual} already deliver strong open-ended report generation and VQA~\cite{antol2015vqa}; however, most models remain optimized for text generation. They answer fluently but provide limited support for the \emph{structured, fine-grained} perception clinicians routinely need—\emph{detection}, \emph{counting}, and \emph{regression}—within a single conversational interface. The FLARE 2D challenge~\cite{FLARE2DTask5}, spanning diverse modalities and task types, makes this gap especially salient and provides a concrete, clinically meaningful testbed.

\begin{figure}
\includegraphics[width=\textwidth]{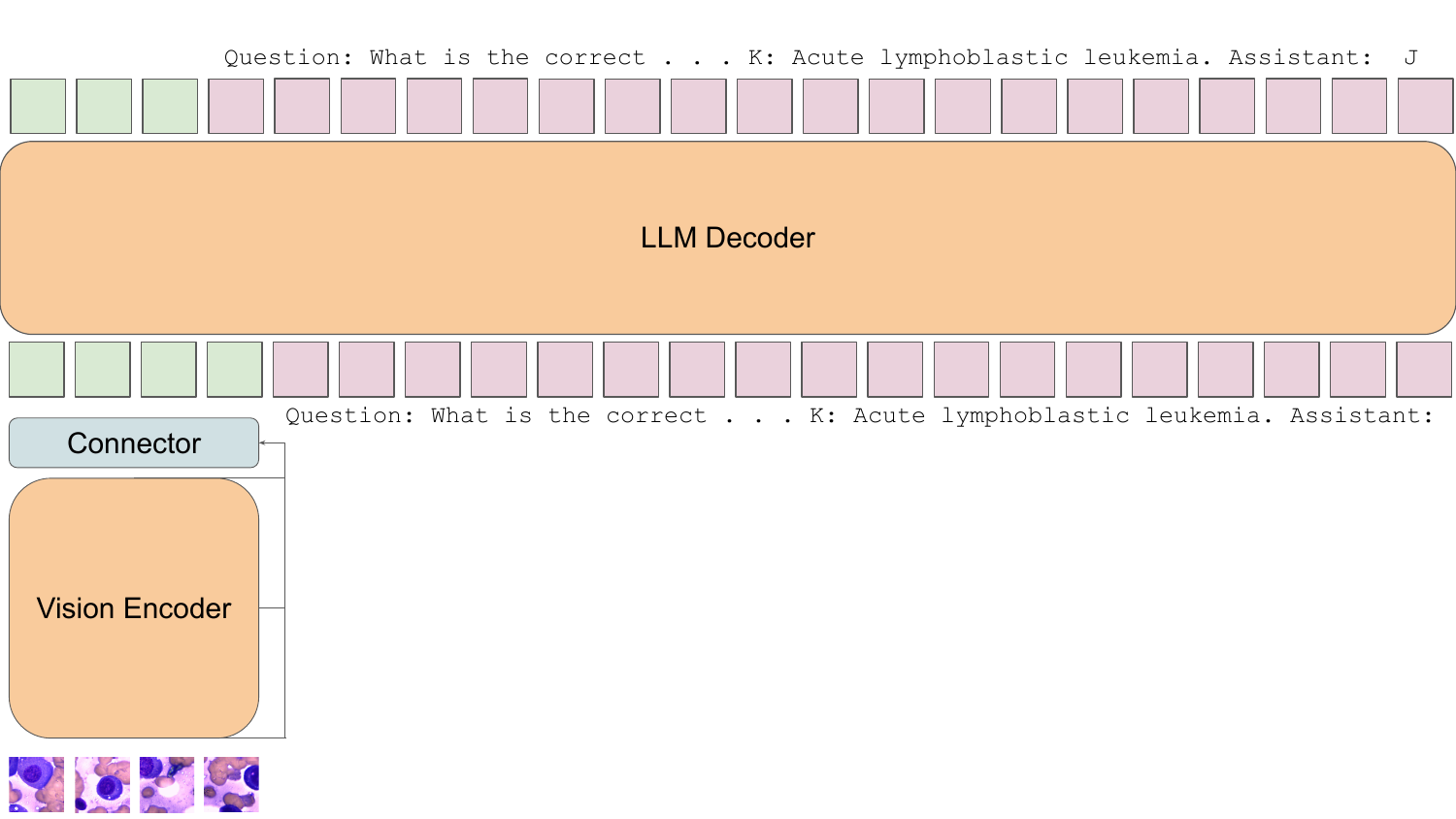}
\caption{\textbf{High-level architecture of \ProposedMethod.}
A Vision Encoder extracts features from medical images; a lightweight dense multi-layer \emph{Connector} fuses low/high-level cues and projects them to the LLM without increasing sequence length. The fused visual tokens (green) are prepended to the text tokens (pink), and a decoder-only LLM performs next-token generation for every task.}
\label{fig:high_level}
\end{figure}

Modern VLMs~\cite{bai2025qwen25vltechnicalreport,liu2023visual,clark2026molmo2} typically couple a pretrained vision encoder to a decoder-only LLM via a lightweight connector and train the stack with next-token prediction over image/video–text corpora~\cite{Nguyen_2026_CVPR,fu2026mme,yu2026dochop}. Variants explore stronger backbones, smarter adapters, and refined alignment/instruction curricula—e.g., PandaGPT aligns multiple perceptual encoders via small projections for unified image/video QA~\cite{su-etal-2023-pandagpt}; Video-LLaMA extends to long-form video with temporal reasoning~\cite{zhang-etal-2023-video}; and Qwen2.5-VL enhances the vision stack with a ViT tower, a \emph{Merger} to consolidate patch embeddings, and improved positional encoding~\cite{bai2025qwen25vltechnicalreport}. In medicine, 2D MLLMs (LLaVA-Med~\cite{NEURIPS2023_5abcdf8e}, HuatuoGPT~\cite{zhang2023huatuogpt}) adapt this recipe through domain alignment and medical instruction tuning, while 3D frameworks pair volumetric encoders with LLMs for CT/MRI dialogue and localization~\cite{NEURIPS2023_5abcdf8e,chen2024huatuogptvisioninjectingmedicalvisual,hamamci2024foundation,xin2025med3dvlm}. Building on these insights, we introduce \ProposedMethod{}, a clinical VLM chatbot that unifies \emph{classification}~\cite{lu2007survey}, \emph{multi-label classification}~\cite{lu2007survey,tsoumakas2007multi}, \emph{detection}~\cite{zhao2019object,nguyen2025revisiting} (textualized outputs), \emph{counting}, \emph{regression}, and free-form \emph{report generation}~\cite{li2018hybrid} under a single next-token interface. The key design choice is a dense, multi-layer \emph{Connector} that aggregates low- and high-level visual features \emph{before} the LLM via channel-wise fusion, enriching tokens without increasing sequence length (Figure~\ref{fig:high_level} and~\ref{fig:connector}).

\paragraph{Contributions:} (i) We create an unified medical VLM chatbot that executes both open-ended \emph{text generation} and \emph{structured} tasks (classification, multi-label classification, detection, counting, regression). (ii) We introduce \ProposedMethod{}, a lightweight dense connector that enriches visual tokens with multi-layer cues, improving fine-grained performance while meeting the challenge’s \emph{fast and efficient} requirement. (iii) We introduce an in-context learning strategy to improve report generation's score.

\section{\ProposedMethod{}}

In this section, we first summarize preliminaries of our method and fix notation (Section~\ref{sec:prelim}), grounding our setup in prior work~\cite{bai2025qwen25vltechnicalreport}. 
Next, we present the proposed \emph{\ProposedMethod{}} architecture that fuses high- and low-level visual features (Section~\ref{sec:connector}). We then detail the \emph{training data} (Section~\ref{sec:training-data}). Finally, we describe our multimodal in-context learning strategies for prompting and adaptation (Section~\ref{sec:icl}).

\subsection{Preliminaries}
\label{sec:prelim}
Our model builds on the Qwen2.5-VL~\cite{bai2025qwen25vltechnicalreport} architecture, which employs a vision encoder $f_\theta$ (ViT with windowed attention) to extract hierarchical visual features from an image or video $x_v$. In Qwen2.5-VL~\cite{bai2025qwen25vltechnicalreport}, a \emph{Merger} module aggregates patch-level embeddings into a sequence of high-level visual tokens. We introduce a \emph{\ProposedMethod{}} module, a custom adapter, that fuses these high-level embeddings with selected low-level features from early encoder layers, enriching the visual representation with fine-grained spatial detail before integration with the language model. Let $\mathcal{V}$ denote the LLM vocabulary and $w_{1:n}=(w_1,\dots,w_n)\in\mathcal{V}^n$ denote a length-$n$ sequence of text tokens. The fused representation $\tilde Z$ is fed into a decoder-only LLM $g_\phi$ as an interleaved sequence $[\tilde{Z}, w_{1:n}]$ for autoregressive text generation.

\subsection{\ProposedMethod{} Architecture}\label{sec:connector}

\begin{figure}[t]
\centering
\includegraphics[width=\textwidth]{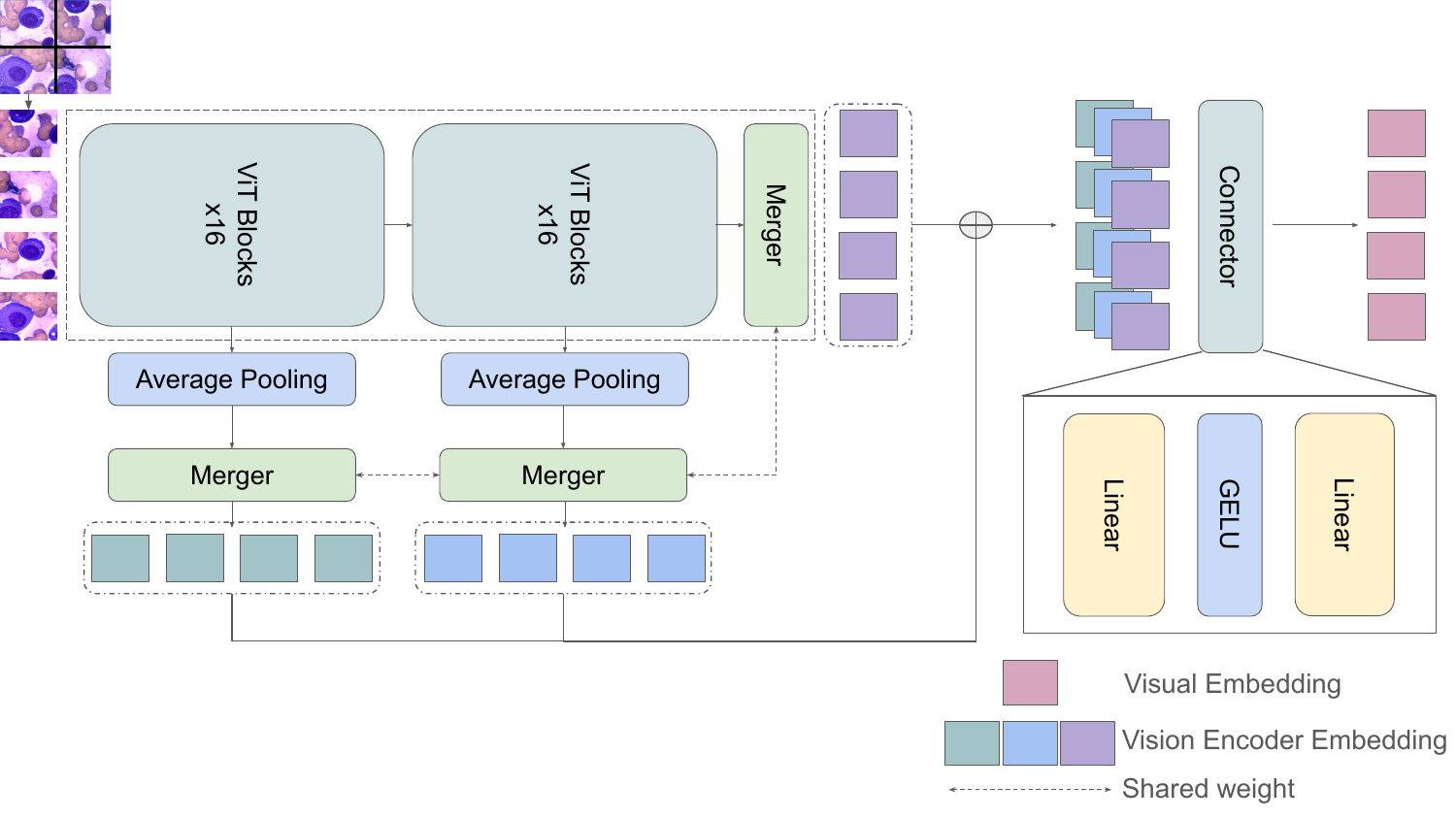}
\caption{\textbf{Dense multi-layer connector in \ProposedMethod{}.}
Early and late ViT features are depth-averaged, aligned via the backbone \emph{Merger}, concatenated channel-wise with the encoder output, and projected by a lightweight MLP to LLM visual tokens.}
\label{fig:connector}
\end{figure}

The vision encoder is a ViT~\cite{dosovitskiy2021an} with \(L{=}32\) blocks. After windowed attention, the hidden state from block \(\ell\) is
\(h^{(\ell)}\in\mathbb{R}^{N\times d_v}\) where \(N\) is number of visual tokens, and \(d_v\) is ViT width.
The encoder’s built-in \emph{Merger} produces output tokens \(Z_0\in\mathbb{R}^{N\times d_m}\).
We reuse that Merger as a mapping \(m_\mu:\mathbb{R}^{N\times d_v}\!\to\!\mathbb{R}^{N\times d_m}\) to align intermediate features.
\(P\) denotes the permutation induced by windowing; \(P^{-1}\) restores the original token order.
The connector MLP is \(c_\psi:\mathbb{R}^{N\times 3d_m}\!\to\!\mathbb{R}^{N\times d_t}\), where \(d_t\) matches the LLM embedding size.

\noindent \textbf{Computation.}
We average features over two depth groups to obtain complementary summaries:
\(G_1{=}\{1,\dots,16\}\) (edge/texture–heavy) and \(G_2{=}\{17,\dots,32\}\) (semantic-heavy),
\[
A_1=\tfrac{1}{|G_1|}\!\sum_{\ell\in G_1} h^{(\ell)},\qquad
A_2=\tfrac{1}{|G_2|}\!\sum_{\ell\in G_2} h^{(\ell)} \in \mathbb{R}^{N\times d_v}.
\]
Each is aligned by the Merger and de-permuted, then concatenated channel-wise with the encoder output and projected:
\[
\;\tilde Z = c_\psi\big(\,[\,Z_0,\; P^{-1}m_\mu(A_1),\; P^{-1}m_\mu(A_2)\,]\,\big)\;\,,
\]
yielding visual tokens \(\tilde Z\in\mathbb{R}^{N\times d_t}\) at the \emph{same} token length \(N\).
These tokens are prepended to the text tokens \(w_{1:n}\) and consumed by the decoder-only LLM for next-token generation as mentioned in ~\ref{sec:prelim}. Detailed visualization of our method is provided in Figure~\ref{fig:connector}.

\subsection{Training Data}\label{sec:training-data}
We use a two-stream mixture for both \emph{alignment} and \emph{instruction tuning}. For alignment, general LLaVA-style image–caption pairs (e.g., LAION~\cite{schuhmann2021laion400mopendatasetclipfiltered,liu2023visual}) are combined with the PubMedVision~\cite{chen2024huatuogptvisioninjectingmedicalvisual} \emph{alignment} split of denoised PubMed image–text pairs~\cite{chen2024huatuogptvisioninjectingmedicalvisual}, yielding \(\sim\)1.2M examples that stabilize open-domain grounding while injecting clinical vocabulary and fine-grained cues. For instruction tuning, we merge LLaVA-Instruct with PubMedVision’s medical instructions~\cite{liu2023visual,chen2024huatuogptvisioninjectingmedicalvisual}, formatting all samples in a unified chat-style template with interleaved image tokens and prompts (\(\sim\)1.3M examples).

As a final stage, we fine-tune on FLARE-2D Task~5~\cite{FLARE2DTask5}, a multimodal medical VQA corpus spanning \emph{8} imaging modalities and \emph{7} task types. We use the official training split (45k question–answer pairs aggregated from 19 datasets) and apply the same chat-style formatting.

\subsection{In-Context Learning}
\label{sec:icl}
Motivated by prior findings that in-context learning (ICL) improves domain use of LLMs without extra training~\cite{many-shot-icl,mm-icl}, we apply ICL to the report-generation task. 
Concretely, we form a support pool $\mathcal{S}$ of $N$ distinct training examples and, at inference, uniformly sample $k$ demonstrations ($k \ll N$). 
Each demonstration is an image–prompt–response triple formatted in our chat template and concatenated before the user query inside a fixed instruction wrapper (Table~\ref{tab:icl_template}). 
No model weights are updated; the LLM conditions on the $k$ demonstrations plus the query to generate the final report.

\begin{table}[h]
    \centering
    \caption{Prompt template with ICL mechanism.}
    \begin{tabular}{|p{200pt}|}
        \hline
        Prompt\\
        \hline
        Instruction: [Task Instruction]\\
        User: [Demonstration 1]\\
        Response: [Answer 1]\\
        ...\\
        User: [Demonstration k]\\
        Response: [Answer k]\\
        User: [Question]\\
        Response: \\
        \hline
    \end{tabular}
    \label{tab:icl_template}
\end{table}

\section{Experimental Results}
In this section, we present: (i) \emph{Hyperparameter setup} (Section~\ref{sec:hyperparams}); (ii) \emph{Evaluation protocol} detailing splits, metrics, prompting templates, and evaluation procedure (Section~\ref{sec:eval}); (iii) \emph{Validation results} across classification, multi-label classification, detection, counting, and regression (Section~\ref{sec:validation}); (iv) \emph{In-context learning} analyses with zero-/few-shot prompting and ablations (Section~\ref{sec:icl-exp}); and (v) \emph{Qualitative results} including case studies and error analyses (Section~\ref{sec:qualitative}).

\subsection{Implementation details}\label{sec:hyperparams}

\subsubsection{Training Protocols.} We adopt a three-stage training pipeline:
(i) \emph{Vision–language alignment} using large-scale image-text corpora to map visual features into the LLM embedding space while stabilizing the language backbone;
(ii) \emph{Visual instruction tuning} on curated multimodal dialogue datasets to elicit robust instruction-following behavior;
(iii) \emph{Domain-specific finetuning} on the FLARE dataset to adapt the model for specialized clinical image understanding tasks.

\subsubsection{Environmental settings.} 
We use AdamW 8-bit~\cite{DettmersLSZ22}, bfloat16 precision, a cosine learning-rate scheduler with warmup ratio = 0.03~\cite{LoshchilovH17}, and max sequence length = 2048. Also, we train our model under 8-bit quantization for memory efficient. Other hyperparameters follow Huggingface's trainer library defaults unless specified in Table~\ref{tab:hyperparams}. Other dependencies and code-related information are available in the codebase provided in the abstract.

\begin{table}
\caption{Key hyperparameters per training stage.}\label{tab:hyperparams}
\begin{center}
\begin{tabular}{|l|l|l|l|}
\hline
 & Alignment & Instruction & FLARE tuning \\
\hline
Epochs & 1 & 1 & 3 \\
Learning rate & 2e-3 & 2e-4 & 2e-4 \\
Per-device batch & 1 & 1 & 1 \\
Number of devices & 8 & 8 & 8 \\
GPU type & A40 & A40 & A40 \\
Training Time & 3 days & 7 days & 1 day\\
Grad. accumulation & 64 & 64 & 16 \\
LoRA (rank, $\alpha$) & -- & (32, 32) & (32, 64) \\
\hline
\end{tabular}
\end{center}
\end{table}

\subsection{Evaluation Protocol}
\label{sec:eval}
We use the FLARE challenge’s two-track validation~\cite{FLARE2DTask5}: \emph{val-hidden} (Codabench) for classification, multi-label, detection, instance detection, regression; and \emph{val-public} (local) for counting and report generation. Metrics are Balanced Accuracy~\cite{brodersen2010balanced} (classification), micro-F1~\cite{grandini2020metrics} (multi-label), F1~\cite{grandini2020metrics} at IoU>0.5 (detection/instance), MAE (regression/counting), and GREEN score~\cite{ostmeier2024green} (report generation).

\subsection{Validation Results}
\label{sec:validation}

\begin{table}
\caption{\textbf{Classification evaluation.} The result is reported on validiation-hidden subset on Codabench. All metrics are reported as fractions in [0,1].}\label{tab:classification}
\begin{center}
\begin{tabular}{|l|l|l|}
\hline
Model &  Classification $\uparrow$ & Multi-label classification $\uparrow$ \\
\hline
4-bit QwenVL 2.5 7B~\cite{qwen25vl-flare2025} & 0.37 & 0.57\\
8-bit QwenVL 2.5 7B~\cite{qwen25vl-flare2025} & 0.36 & 0.56\\
16-bit QwenVL 2.5 7B~\cite{qwen25vl-flare2025} & 0.35 & 0.55\\
\hline
4-bit \ProposedMethod{} 7B (Ours) & 0.51 & 0.49\\
8-bit \ProposedMethod{} 7B (Ours) & 0.46 & 0.53 \\
16-bit \ProposedMethod{} 7B (Ours) & 0.50 & 0.54\\
\hline
\end{tabular}
\end{center}
\end{table}

\begin{table}
\caption{\textbf{Detection evaluation.} The result is reported on validiation-hidden subset on Codabench. All metrics are reported as fractions in [0,1].}\label{sec:detection}
\begin{center}
\begin{tabular}{|l|l|l|}
\hline
Model &  Detection $\uparrow$ & Instance Detection $\uparrow$ \\
\hline
4-bit QwenVL 2.5 7B~\cite{qwen25vl-flare2025} & 0.51 & 0\\
8-bit QwenVL 2.5 7B~\cite{qwen25vl-flare2025} & 0.55 & 0\\
16-bit QwenVL 2.5 7B~\cite{qwen25vl-flare2025} & 0.53 & 0\\
\hline
4-bit \ProposedMethod{} 7B (Ours) & 0.76 & 0\\
8-bit \ProposedMethod{} 7B (Ours) & 0.85 & 0 \\
16-bit \ProposedMethod{} 7B (Ours) & 0.81 & 0\\
\hline
\end{tabular}
\end{center}
\end{table}

\paragraph{Classification evaluation.}
On the Codabench validation-hidden split (Table~\ref{tab:classification}), \ProposedMethod{} surpasses the Qwen2.5-VL baseline in \emph{single-label classification} at all bit-widths: +0.14 at 4-bit quantization (from 0.37 to 0.51), +0.1 at 8-bit quantization (from 0.36 to 0.46), and +0.15 at 16-bit quantization (from 0.35 to 0.50). For \emph{multi-label classification}, as the Micro-averaged F1 increases with precision - 0.49 (4-bit quantization), 0.53 (8-bit), and 0.54 (16-bit) - \ProposedMethod{} approaches the baseline (0.57), reducing the gap from $-0.08$ to $-0.03$. Overall, 8-bit quantization offers a strong efficiency–accuracy trade-off (large single-label gains with competitive multi-label scores), while 16-bit quantization maximizes multi-label performance.

\paragraph{Detection evaluation.}
Using F1 as the metric, \ProposedMethod{} markedly outperforms the Qwen2.5-VL baseline on detection in Table~\ref{sec:detection}: 0.76 at 4-bit quantization (\(+0.25\), from 0.51 to 0.76), 0.85 at 8-bit quantization (\(+0.3\), from 0.55 to 0.85), and 0.81 at 16-bit quantization (\(+0.28\), from 0.53 to 0.81). The best F1 is achieved at 8-bit quantization, suggesting an effective efficiency–accuracy sweet spot. \emph{Instance detection} F1 is 0 across all settings, indicating that our current textualized detection output does not satisfy the instance-level scoring protocol; enabling structured instance outputs is left for future work.

\begin{table}
\caption{\textbf{Regression, Counting, and Report Generation evaluation.} The result of Regression is reported on validation-hidden subset on Codabench. The result of Counting and Report Generation is reported on validation-public subset. Report Generation is reported as decimal number in [0, 100].}\label{tab:reg}
\begin{center}
\begin{tabular}{|l|l|l|l|}
\hline
Model &  Regression $\downarrow$ & Counting $\downarrow$ & Report Generation $\uparrow$ \\
\hline
4-bit QwenVL 2.5 7B~\cite{qwen25vl-flare2025} & 15.51 & 287.3 & 65.74 \\
8-bit QwenVL 2.5 7B~\cite{qwen25vl-flare2025} & 15.19 & 276.77 & 55.77  \\
16-bit QwenVL 2.5 7B~\cite{qwen25vl-flare2025} & 15.43 & 268.24 & 74.78  \\
\hline
4-bit \ProposedMethod{} 7B (Ours) & 20.91 & 284.44 & 76.01\\
8-bit \ProposedMethod{} 7B (Ours) & 21.13 & 275.81 & 74.05\\
16-bit \ProposedMethod{} 7B (Ours) & 19.04 & 266.7 & 74.93\\
\hline
\end{tabular}
\end{center}
\end{table}

\paragraph{Regression, counting, and report generation.}
From Table~\ref{tab:reg}, \emph{counting} decreases at all settings: from 287.30 to 284.44 ($-2.86$) at 4-bit quantization, from 276.77 to 275.81 ($-0.96$) at 8-bit quantization, and from 268.24 to 266.70 ($-1.54$) at 16-bit quantization, with the best MAE at 16-bit. \emph{Report generation} increases at 4-bit quantization from 65.74 to 76.01 ($+10.27$), at 8-bit quantization from 55.77 to 74.05 ($+18.28$), and at 16-bit quantization from 74.78 to 74.93 ($+0.15$), peaking at 4-bit. In contrast, \emph{regression} increases—at 4-bit quantization from 15.51 to 20.91 ($+5.40$), at 8-bit quantization from 15.19 to 21.13 ($+5.94$), and at 16-bit quantization from 15.43 to 19.04 ($+3.61$)—highlighting a remaining gap for purely numeric targets.

\subsection{In-Context Learning Results}
\label{sec:icl-exp}
\begin{figure}[h]
    \centering
    \renewcommand{\arraystretch}{0.7}
    \begin{tabular}{cc}
        \includegraphics[width=0.5\textwidth, height=0.35\textwidth]{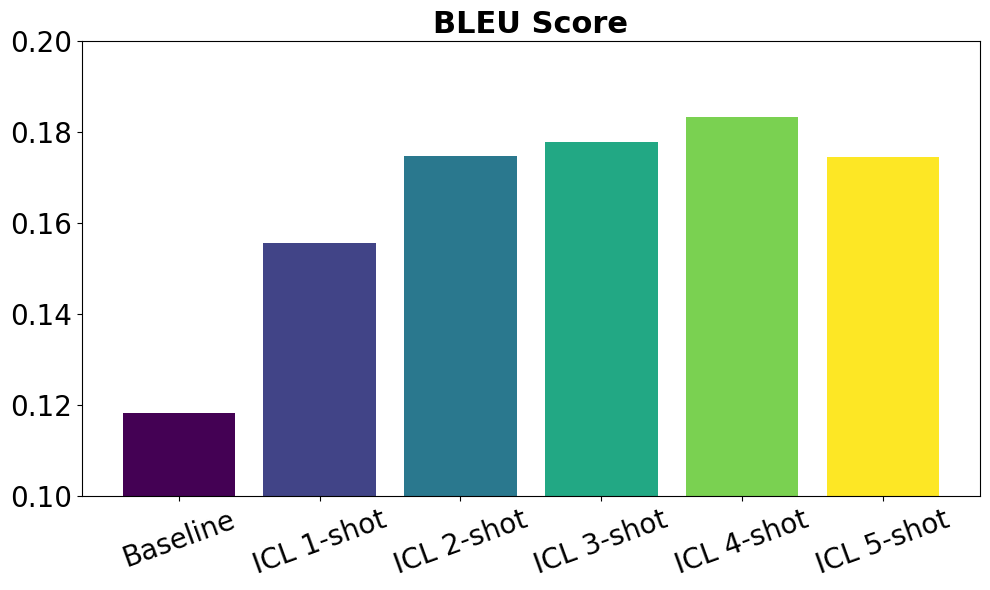} & 
        \includegraphics[width=0.5\textwidth, height=0.35\textwidth]{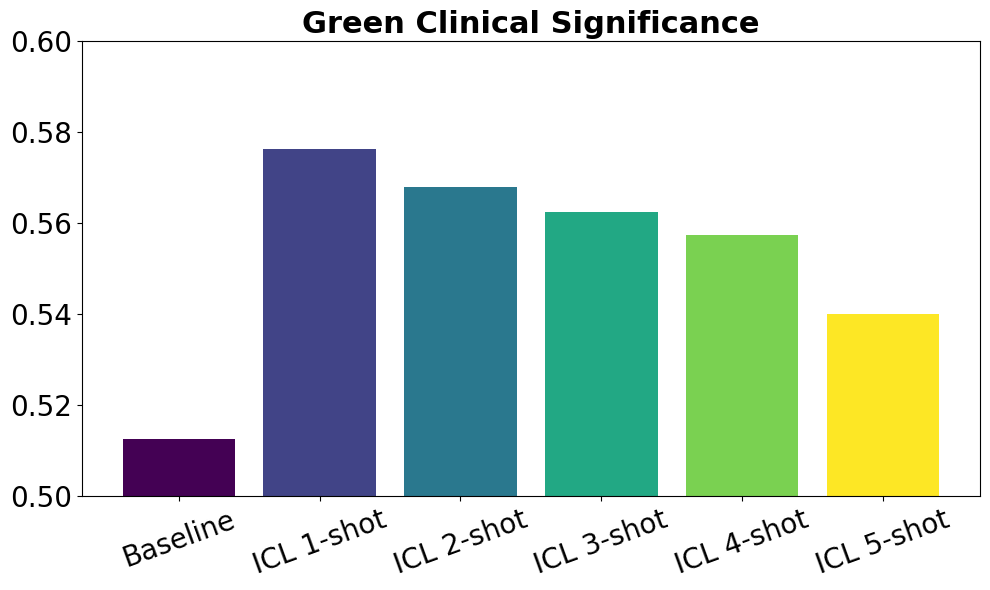}\\
        \includegraphics[width=0.5\textwidth, height=0.35\textwidth]{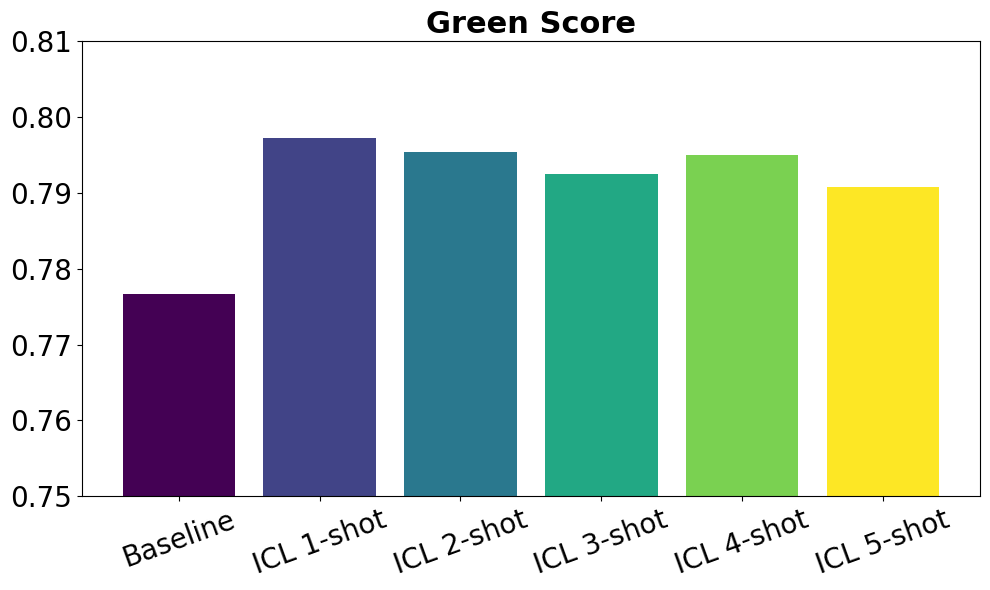} & 
        \includegraphics[width=0.5\textwidth, height=0.35\textwidth]{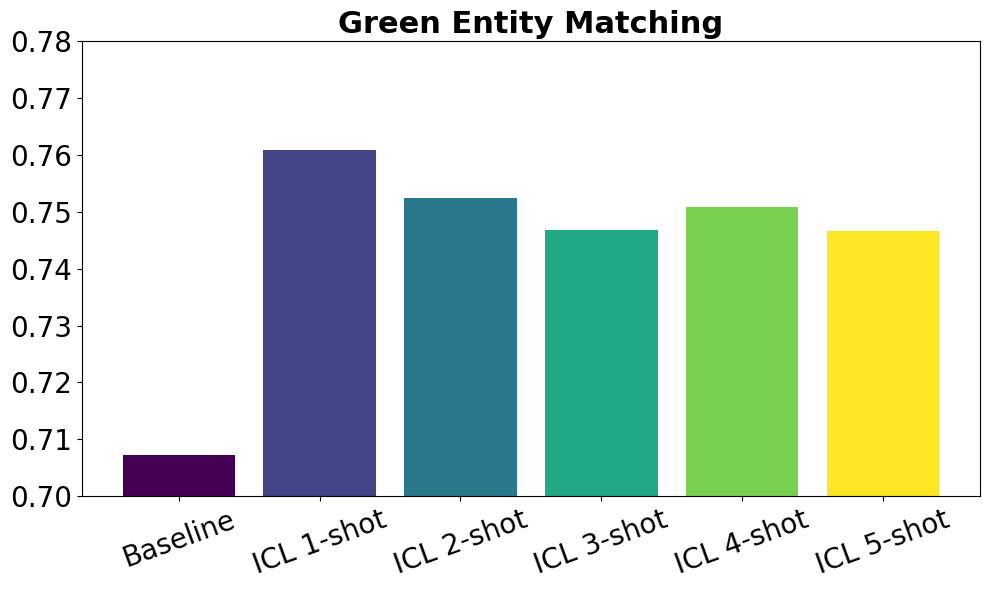}\\
    \end{tabular}
    \caption{Effect of in-context learning (ICL) on \ProposedMethod{} for the report-generation task: BLEU, GREEN Clinical Significance, GREEN Entity Matching, and overall GREEN Score on the validation-public split. Incorporating $k$ demonstrations consistently improves performance over no-ICL prompts.}
    \label{fig:icl_performance}
\end{figure}

Figure~\ref{fig:icl_performance} summarizes the effect of adding $k$ in-context demonstrations for report generation on the validation-public split. 
Across all four metrics—BLEU~\cite{tantug2008bleu+}, GREEN Clinical Significance, GREEN Entity Matching, and overall GREEN Score—ICL consistently improves performance compared with prompts without demonstrations. 
These gains indicate that brief, in-domain exemplars help the model follow radiology style and terminology more faithfully, complementing our connector-based improvements without additional finetuning. Due to compute/memory limits on the competition testing server and Docker submission constraints, ICL was not enabled in our container; the results in Figure~\ref{fig:icl_performance} are from offline runs and are shown as a potential future direction.

\begin{figure}[h]
    \centering
    \includegraphics[width=\textwidth]{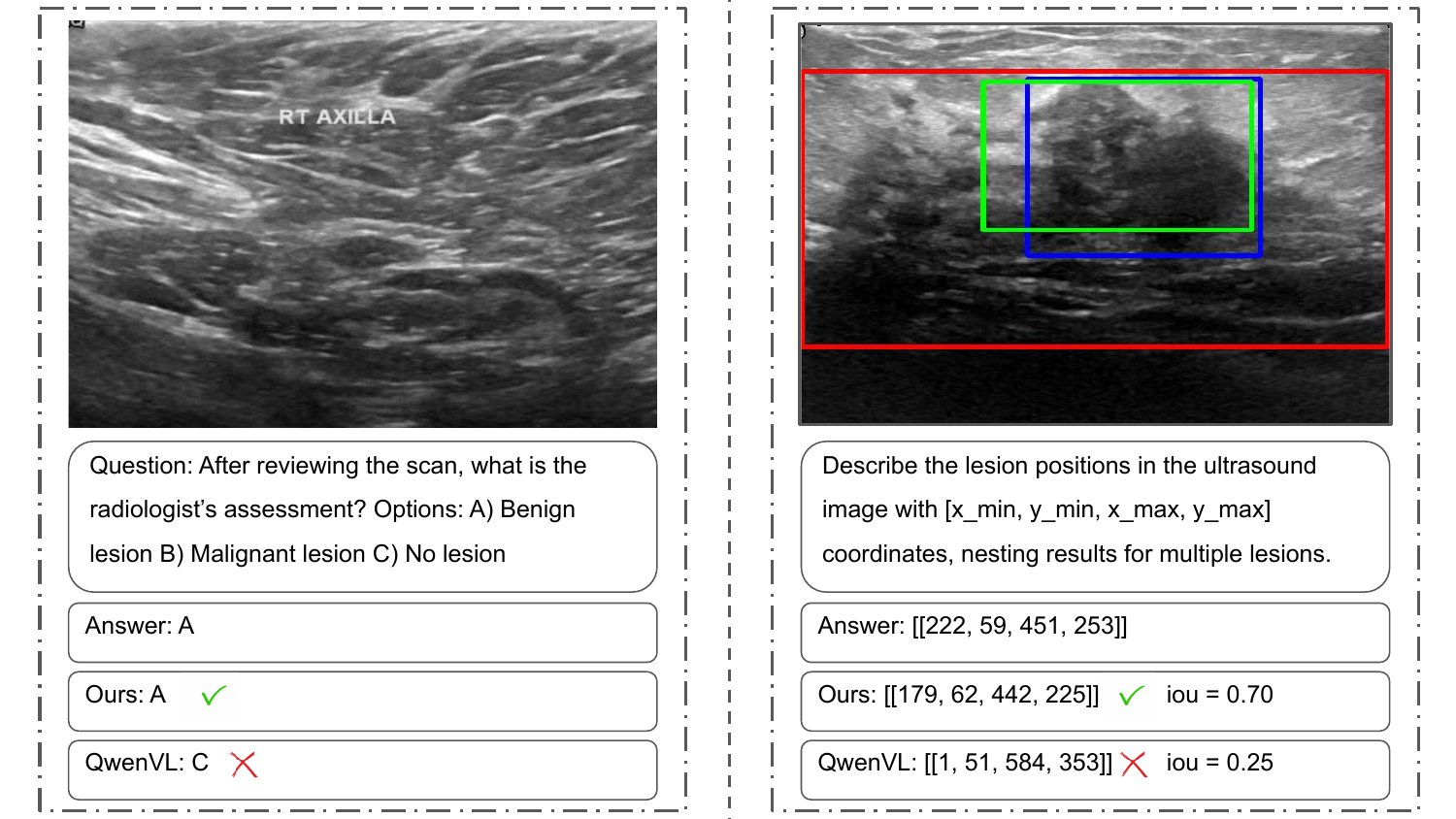}
    \caption{\textbf{Qualitative examples on Ultrasound images.} Left: multiple-choice classification task. Right: lesion detection task where \textcolor{OliveGreen}{green} is our detection, \textcolor{blue}{blue}is ground truth, and \textcolor{red}{red} is QwenVL's~\cite{qwen25vl-flare2025} detection.}
    \label{fig:qualitative}
\end{figure}

\subsection{Qualitative Result}
\label{sec:qualitative}
\noindent \textbf{Qualitative results.}
Figure~\ref{fig:qualitative} showcases ultrasound images for two tasks. 
\emph{Left (classification):} \ProposedMethod{} selects the correct option (A), whereas QwenVL 2.5~\cite{qwen25vl-flare2025} predicts C. 
\emph{Right (detection):} our textualized box (green) aligns more closely with the ground truth (blue), achieving IoU\,=\,0.70, while QwenVL 2.5's~\cite{qwen25vl-flare2025} box (red) attains IoU\,=\,0.25. 
These examples underscore \ProposedMethod{}’s stronger recognition and localization on ultrasound imagery.

\section{Limitation and future work}
Despite unifying open-ended and structured tasks, the system has several gaps: instance-level prediction is not realized because textualized detection cannot satisfy instance scoring; numeric grounding is weak, with regression lagging and counting improvements modest; performance is sensitive to precision/quantization and training remains compute-intensive; and safety/consistency checks remain limited relative to clinical requirements. In addition, the in-context learning (ICL) strategy substantially increases prompt length and inference memory/latency. Therefore, an in-depth, task-specific study of demonstration budgeting is needed to balance accuracy, cost, and latency. Future work will need to explore stronger connector architectures that more effectively fuse multi-layer visual cues, improving structured prediction while preserving efficiency.

\section{Conclusion}

\ProposedMethod{} advances medical VLMs by unifying open-ended report generation with structured perception—classification, multi-label classification, detection, counting, and regression—within a single next-token interface. Built on Qwen2.5-VL with a lightweight dense multi-layer connector, the system improves fine-grained recognition while preserving efficiency and delivers consistent gains on FLARE 2D tasks, particularly for classification and detection. Nonetheless, gaps remain in instance-level prediction and numeric grounding. Future work will need to explore stronger connector architectures that fuse multi-layer visual cues more effectively to enhance structured prediction while maintaining speed and simplicity.

\subsubsection{Acknowledgements} This research is partially supported by the following NIH grants: R01-HL171376 and U01-CA268808. The authors of this paper declare that the proposed solution is fully automatic without any manual intervention. We thank all data owners and contributors for making the data publicly available and CodaLab~\cite{codabench} for hosting the challenge platform. We also thank AI VIETNAM for supporting the GPU resource for conducting the experiments.

\section*{Disclosure of Interests}
The authors declare no competing interests.

\begin{comment}  %% removed for anonymized MICCAI 2025 submission.
    
    % The following acknowledgement and disclaimer sections should be removed for the double-blind review process.  
    % If and when your paper is accepted, reinsert the acknowledgement and the disclaimer clause in your final camera-ready version.

\begin{credits}
\subsubsection{\ackname} A bold run-in heading in small font size at the end of the paper is
used for general acknowledgments, for example: This study was funded
by X (grant number Y).

\subsubsection{\discintname}
It is now necessary to declare any competing interests or to specifically
state that the authors have no competing interests. Please place the
statement with a bold run-in heading in small font size beneath the
(optional) acknowledgments\footnote{If EquinOCS, our proceedings submission
system, is used, then the disclaimer can be provided directly in the system.},
for example: The authors have no competing interests to declare that are
relevant to the content of this article. Or: Author A has received research
grants from Company W. Author B has received a speaker honorarium from
Company X and owns stock in Company Y. Author C is a member of committee Z.
\end{credits}

\end{comment}
%
% ---- Bibliography ----
%
% BibTeX users should specify bibliography style 'splncs04'.
% References will then be sorted and formatted in the correct style.
%
\bibliographystyle{splncs04}
\bibliography{mybibliography}

@misc{bai2025qwen25vltechnicalreport,
      title={Qwen2.5-VL Technical Report}, 
      author={Shuai Bai and Keqin Chen and Xuejing Liu and Jialin Wang and Wenbin Ge and Sibo Song and Kai Dang and Peng Wang and Shijie Wang and Jun Tang and Humen Zhong and Yuanzhi Zhu and Mingkun Yang and Zhaohai Li and Jianqiang Wan and Pengfei Wang and Wei Ding and Zheren Fu and Yiheng Xu and Jiabo Ye and Xi Zhang and Tianbao Xie and Zesen Cheng and Hang Zhang and Zhibo Yang and Haiyang Xu and Junyang Lin},
      year={2025},
      eprint={2502.13923},
      archivePrefix={arXiv},
      primaryClass={cs.CV},
      url={https://arxiv.org/abs/2502.13923}, 
}

@inproceedings{su-etal-2023-pandagpt,
    title = "{P}anda{GPT}: One Model To Instruction-Follow Them All",
    author = "Su, Yixuan  and
      Lan, Tian  and
      Li, Huayang  and
      Xu, Jialu  and
      Wang, Yan  and
      Cai, Deng",
    booktitle = "Proceedings of the 1st Workshop on Taming Large Language Models: Controllability in the era of Interactive Assistants!",
    month = sep,
    year = "2023",
    address = "Prague, Czech Republic",
    publisher = "Association for Computational Linguistics",
    url = "https://aclanthology.org/2023.tllm-1.2/",
    pages = "11--23"
}

@inproceedings{LoshchilovH17,
  author       = {Ilya Loshchilov and
                  Frank Hutter},
  title        = {{SGDR:} Stochastic Gradient Descent with Warm Restarts},
  booktitle    = {5th International Conference on Learning Representations, {ICLR}},
  year         = {2017},
  url          = {https://openreview.net/forum?id=Skq89Scxx},
  bibsource    = {dblp computer science bibliography, https://dblp.org}
}

@inproceedings{DettmersLSZ22,
  author       = {Tim Dettmers and
                  Mike Lewis and
                  Sam Shleifer and
                  Luke Zettlemoyer},
  title        = {8-bit Optimizers via Block-wise Quantization},
  booktitle    = {The Tenth International Conference on Learning Representations, {ICLR}},
  year         = {2022},
  url          = {https://openreview.net/forum?id=shpkpVXzo3h},
  bibsource    = {dblp computer science bibliography, https://dblp.org}
}

@misc{FLARE2DTask5,
      title={Fast, Low-resource, Accurate, Robust, and Effectual Medical Image Analysis (FLARE) 2025}, 
      author={Jun Ma and Song Gu and Bo Wang},
      year={Mar. 27, 2025},
      archivePrefix={Zenodo},
      url={doi: 10.5281/zenodo.15094799}, 
}

@inproceedings{antol2015vqa,
  title={Vqa: Visual question answering},
  author={Antol, Stanislaw and Agrawal, Aishwarya and Lu, Jiasen and Mitchell, Margaret and Batra, Dhruv and Zitnick, C Lawrence and Parikh, Devi},
  booktitle={Proceedings of the IEEE international conference on computer vision},
  pages={2425--2433},
  year={2015}
}

@inproceedings{
dosovitskiy2021an,
title={An Image is Worth 16x16 Words: Transformers for Image Recognition at Scale},
author={Alexey Dosovitskiy and Lucas Beyer and Alexander Kolesnikov and Dirk Weissenborn and Xiaohua Zhai and Thomas Unterthiner and Mostafa Dehghani and Matthias Minderer and Georg Heigold and Sylvain Gelly and Jakob Uszkoreit and Neil Houlsby},
booktitle={International Conference on Learning Representations},
year={2021},
url={https://openreview.net/forum?id=YicbFdNTTy}
}

@inproceedings{
yu2026dochop,
title={DocHop: Benchmarking Out-of-domain Multi-hop Reasoning in Information-Dense Documents},
author={Zhuoran Yu and Le Thien Phuc Nguyen and Jaden Park and Xinyi Gu and Zexue He and Soochahn Lee and Rogerio Feris and Yong Jae Lee},
booktitle={Forty-third International Conference on Machine Learning},
year={2026},
url={https://openreview.net/forum?id=PQFkScoGqz}
}

@article{fu2026mme,
  title={Mme: A comprehensive evaluation benchmark for multimodal large language models},
  author={Fu, Chaoyou and Chen, Peixian and Shen, Yunhang and Qin, Yulei and Zhang, Mengdan and Lin, Xu and Yang, Jinrui and Zheng, Xiawu and Li, Ke and Sun, Xing and others},
  journal={Advances in Neural Information Processing Systems},
  volume={38},
  year={2026}
}

@InProceedings{Nguyen_2026_CVPR,
    author    = {Nguyen, Le Thien Phuc and Yu, Zhuoran and Hang, Samuel Low Yu and An, Subin and Lee, Jeongik and Ban, Yohan and Chung, SeungEun and Nguyen, Thanh-Huy and Maeng, JuWan and Lee, Soochahn and Lee, Yong Jae},
    title     = {See, Hear, and Understand: Benchmarking Audiovisual Human Speech Understanding in Multimodal Large Language Models},
    booktitle = {Proceedings of the IEEE/CVF Conference on Computer Vision and Pattern Recognition (CVPR) Findings},
    month     = {June},
    year      = {2026},
    pages     = {2272-2283}
}

@inproceedings{tantug2008bleu+,
  title={BLEU+: a Tool for Fine-Grained BLEU Computation.},
  author={Tantug, A C{\"u}neyd and Oflazer, Kemal and El-Kahlout, Ilknur Durgar},
  booktitle={LREC},
  year={2008}
}

@article{nguyen2025revisiting,
  title={Revisiting Active Speaker Detection: An In-the-Wild Benchmark for Generalization and Robustness},
  author={Nguyen, Le Thien Phuc and Yu, Zhuoran and Cao, Khoa Quang Nhat and Guo, Yuwei and Pham, Tu Ho Manh and Nguyen, Tuan Tai and Vo, Toan Ngo Duc and Poon, Lucas and Nguyen, Tuan Khai and Lee, Soochahn and others},
  journal={arXiv preprint arXiv:2505.21954},
  year={2025}
}

@article{li2018hybrid,
  title={Hybrid retrieval-generation reinforced agent for medical image report generation},
  author={Li, Yuan and Liang, Xiaodan and Hu, Zhiting and Xing, Eric P},
  journal={Advances in neural information processing systems},
  volume={31},
  year={2018}
}

@article{zhao2019object,
  title={Object detection with deep learning: A review},
  author={Zhao, Zhong-Qiu and Zheng, Peng and Xu, Shou-tao and Wu, Xindong},
  journal={IEEE transactions on neural networks and learning systems},
  volume={30},
  number={11},
  pages={3212--3232},
  year={2019},
  publisher={IEEE}
}

@article{lu2007survey,
  title={A survey of image classification methods and techniques for improving classification performance},
  author={Lu, Dengsheng and Weng, Qihao},
  journal={International journal of Remote sensing},
  volume={28},
  number={5},
  pages={823--870},
  year={2007},
  publisher={Taylor \& Francis}
}

@article{tsoumakas2007multi,
  title={Multi-label classification: An overview},
  author={Tsoumakas, Grigorios and Katakis, Ioannis},
  journal={International Journal of Data Warehousing and Mining (IJDWM)},
  volume={3},
  number={3},
  pages={1--13},
  year={2007},
  publisher={IGI Global Scientific Publishing}
}

@article{grandini2020metrics,
  title={Metrics for multi-class classification: an overview},
  author={Grandini, Margherita and Bagli, Enrico and Visani, Giorgio},
  journal={arXiv preprint arXiv:2008.05756},
  year={2020}
}

@inproceedings{clark2026molmo2,
  title={Molmo2: Open weights and data for vision-language models with video understanding and grounding},
  author={Clark, Christopher and Zhang, Jieyu and Ma, Zixian and Park, Jae Sung and Tripathi, Rohun and Lee, Sangho and Salehi, Mohammadreza and Ren, Jason and Kim, Chris Dongjoo and Yang, Yinuo and others},
  booktitle={Proceedings of the IEEE/CVF Conference on Computer Vision and Pattern Recognition},
  pages={28652--28668},
  year={2026}
}

@inproceedings{brodersen2010balanced,
  title={The balanced accuracy and its posterior distribution},
  author={Brodersen, Kay Henning and Ong, Cheng Soon and Stephan, Klaas Enno and Buhmann, Joachim M},
  booktitle={2010 20th international conference on pattern recognition},
  pages={3121--3124},
  year={2010},
  organization={IEEE}
}

@inproceedings{ostmeier2024green,
  title={Green: Generative radiology report evaluation and error notation},
  author={Ostmeier, Sophie and Xu, Justin and Chen, Zhihong and Varma, Maya and Blankemeier, Louis and Bluethgen, Christian and Md, Arne Edward Michalson and Moseley, Michael and Langlotz, Curtis and Chaudhari, Akshay S and others},
  booktitle={Findings of the association for computational linguistics: EMNLP 2024},
  pages={374--390},
  year={2024}
}

@inproceedings{zhang2023huatuogpt,
  title={Huatuogpt, towards taming language model to be a doctor},
  author={Zhang, Hongbo and Chen, Junying and Jiang, Feng and Yu, Fei and Chen, Zhihong and Chen, Guiming and Li, Jianquan and Wu, Xiangbo and Zhiyi, Zhang and Xiao, Qingying and others},
  booktitle={Findings of the association for computational linguistics: EMNLP 2023},
  pages={10859--10885},
  year={2023}
}

@misc{schuhmann2021laion400mopendatasetclipfiltered,
      title={LAION-400M: Open Dataset of CLIP-Filtered 400 Million Image-Text Pairs}, 
      author={Christoph Schuhmann and Richard Vencu and Romain Beaumont and Robert Kaczmarczyk and Clayton Mullis and Aarush Katta and Theo Coombes and Jenia Jitsev and Aran Komatsuzaki},
      year={2021},
      eprint={2111.02114},
      archivePrefix={arXiv},
      primaryClass={cs.CV},
      url={https://arxiv.org/abs/2111.02114}, 
}

@article{liu2023visual,
  title={Visual instruction tuning},
  author={Liu, Haotian and Li, Chunyuan and Wu, Qingyang and Lee, Yong Jae},
  journal={Advances in neural information processing systems},
  volume={36},
  pages={34892--34916},
  year={2023}
}

@misc{chen2024huatuogptvisioninjectingmedicalvisual,
      title={HuatuoGPT-Vision, Towards Injecting Medical Visual Knowledge into Multimodal LLMs at Scale}, 
      author={Junying Chen and Chi Gui and Ruyi Ouyang and Anningzhe Gao and Shunian Chen and Guiming Hardy Chen and Xidong Wang and Ruifei Zhang and Zhenyang Cai and Ke Ji and Guangjun Yu and Xiang Wan and Benyou Wang},
      year={2024},
      eprint={2406.19280},
      archivePrefix={arXiv},
      primaryClass={cs.CV},
      url={https://arxiv.org/abs/2406.19280}, 
}

@article{xin2025med3dvlm,
  title={Med3DVLM: An Efficient Vision-Language Model for 3D Medical Image Analysis},
  author={Xin, Yu and Ates, Gorkem Can and Gong, Kuang and Shao, Wei},
  journal={arXiv preprint arXiv:2503.20047},
  year={2025}
}

@misc{hamamci2024foundation,
      title={Developing Generalist Foundation Models from a Multimodal Dataset for 3D Computed Tomography}, 
      author={Ibrahim Ethem Hamamci and Sezgin Er and Furkan Almas and Ayse Gulnihan Simsek and Sevval Nil Esirgun and Irem Dogan and Muhammed Furkan Dasdelen and Omer Faruk Durugol and Bastian Wittmann and Tamaz Amiranashvili and Enis Simsar and Mehmet Simsar and Emine Bensu Erdemir and Abdullah Alanbay and Anjany Sekuboyina and Berkan Lafci and Christian Bluethgen and Mehmet Kemal Ozdemir and Bjoern Menze},
      year={2024},
      eprint={2403.17834},
      archivePrefix={arXiv},
      primaryClass={cs.CV},
      url={https://arxiv.org/abs/2403.17834}, 
}

@misc{qwen25vl-flare2025,
  title={QWen2.5VL Fine-tuned for FLARE 2025 Medical Image Analysis},
  author={Shuolin Yin},
  year={2025},
  publisher={Hugging Face},
  url={https://huggingface.co/leoyinn/qwen2.5vl-flare2025}
}

@inproceedings{NEURIPS2023_5abcdf8e,
 author = {Li, Chunyuan and Wong, Cliff and Zhang, Sheng and Usuyama, Naoto and Liu, Haotian and Yang, Jianwei and Naumann, Tristan and Poon, Hoifung and Gao, Jianfeng},
 booktitle = {Advances in Neural Information Processing Systems},
 editor = {A. Oh and T. Naumann and A. Globerson and K. Saenko and M. Hardt and S. Levine},
 pages = {28541--28564},
 publisher = {Curran Associates, Inc.},
 title = {LLaVA-Med: Training a Large Language-and-Vision Assistant for Biomedicine in One Day},
 volume = {36},
 year = {2023}
}

@article{codabench,
    title = {Codabench: Flexible, easy-to-use, and reproducible meta-benchmark platform},
    author = {Zhen Xu and Sergio Escalera and Adrien Pavão and Magali Richard and Wei-Wei Tu and Quanming Yao and Huan Zhao and Isabelle Guyon},
    journal = {Patterns},
    volume = {3},
    number = {7},
    pages = {100543},
    year = {2022}
}

@inproceedings{zhang-etal-2023-video,
    title = "Video-{LL}a{MA}: An Instruction-tuned Audio-Visual Language Model for Video Understanding",
    author = "Zhang, Hang  and
      Li, Xin  and
      Bing, Lidong",
    booktitle = "Proceedings of the 2023 Conference on Empirical Methods in Natural Language Processing: System Demonstrations",
    month = dec,
    year = "2023",
    address = "Singapore",
    publisher = "Association for Computational Linguistics",
    url = "https://aclanthology.org/2023.emnlp-demo.49/",
    doi = "10.18653/v1/2023.emnlp-demo.49",
    pages = "543--553"
}

@inproceedings{
many-shot-icl,
title={Many-shot In-Context Learning},
author={Rishabh Agarwal and Avi Singh and Lei M Zhang and Bernd Bohnet and Luis Rosias and Stephanie C.Y. Chan and Biao Zhang and Aleksandra Faust and Hugo Larochelle},
booktitle={ICML 2024 Workshop on In-Context Learning},
year={2024},
url={https://openreview.net/forum?id=goi7DFHlqS}
}

@inproceedings{
mm-icl,
title={What Factors Affect Multi-Modal In-Context Learning? An In-Depth Exploration},
author={Libo Qin and Qiguang Chen and Hao Fei and Zhi Chen and Min Li and Wanxiang Che},
booktitle={The Thirty-eighth Annual Conference on Neural Information Processing Systems},
year={2024},
url={https://openreview.net/forum?id=REVdYKGcfb}
}

\newpage
% Please add the following required packages to your document preamble:
% \usepackage[normalem]{ulem}
% \useunder{\uline}{\ul}{}
\begin{table}[!htbp]
\caption{Checklist Table. Please fill out this checklist table in the answer column.}
\centering
\begin{tabular}{ll}
\hline
Requirements                                                                                                                    & Answer        \\ \hline
A meaningful title                                                                                                              & Yes        \\ \hline
The number of authors ($\leq$6)                                                                                                             & 6        \\ \hline
Author affiliations and ORCID                                                                                           & Yes        \\ \hline
Corresponding author email is presented                                                                                                  & Yes        \\ \hline
Validation scores are presented in the abstract                                                                                 & Yes        \\ \hline
\begin{tabular}[c]{@{}l@{}}Introduction includes at least three parts: \\ background, related work, and motivation\end{tabular} & Yes        \\ \hline
A pipeline/network figure is provided                                                                                           & Figure~\ref{fig:high_level},~\ref{fig:connector} \\ \hline
Pre-processing (Training data)                                                                                                                  & Page 4   \\ \hline

Strategies to improve model inference                                                                                           & Page 4   \\ \hline

The dataset and evaluation metric section are presented                                                                              & Page 5   \\ \hline
Environment setting table is provided                                                                                           & Table~\ref{tab:hyperparams}  \\ \hline
Training protocol table is provided                                                                                             & Page 5  \\ \hline

Visualized example is provided                                                                                     & Figure~\ref{fig:qualitative} \\ \hline
Limitation and future work are presented                                                                                        & Yes        \\ \hline
Reference format is consistent.  & Yes        \\ \hline

\end{tabular}
\end{table}
\end{document}